\documentclass[11pt]{article}

\usepackage[final]{acl}

\usepackage{times}
\usepackage{latexsym}
\usepackage{tabularx}
\usepackage{booktabs}
\usepackage{amsmath}

\usepackage{graphicx}
\usepackage{listings}
\usepackage{caption}
\usepackage{xcolor} 

\usepackage[T1]{fontenc}

\usepackage[utf8]{inputenc}

\usepackage{microtype}

\usepackage{inconsolata}

\usepackage{graphicx}

\title{MasonNLP at MEDIQA-OE 2025: Assessing Large Language Models for Structured Medical Order Extraction}

\author{A H M Rezaul Karim \\
  George Mason University, VA, USA\\
  \texttt{akarim9@gmu.edu} \\\And
  \"Ozlem Uzuner\\
  George Mason University, VA, USA\\
  \texttt{ouzuner@gmu.edu} \\}

\begin{document}
\maketitle
\begin{abstract}
Medical order extraction is essential for structuring actionable clinical information, supporting decision-making, and enabling downstream applications such as documentation and workflow automation. Orders may be embedded in diverse sources, including electronic health records, discharge summaries, and multi-turn doctor–patient dialogues, and can span categories such as medications, laboratory tests, imaging studies, and follow-up actions. The MEDIQA-OE 2025 shared task focuses on extracting structured medical orders from extended conversational transcripts, requiring the identification of order type, description, reason, and provenance. We present the MasonNLP submission, which ranked 5\textsuperscript{th} among 17 participating teams with 105 total submissions. Our approach uses a general-purpose, instruction-tuned LLaMA-4 17B model without domain-specific fine-tuning, guided by a single in-context example. This few-shot configuration achieved an average $F_1$ score of 37.76, with notable improvements in reason and provenance accuracy. These results demonstrate that large, non-domain-specific LLMs, when paired with effective prompt engineering, can serve as strong, scalable baselines for specialized clinical NLP tasks. \footnote{Implementation can be found here: \url{ https://github.com/AHMRezaul/MEDIQA-OE-2025}}
\end{abstract}

\section{Introduction}


Clinical free-text notes in electronic health records (EHRs) contain essential information such as diagnoses, medications, procedures, and treatment plans  \cite{wang2018clinical, demner2009can}. Extracting structured medical orders, including medications, labs, imaging, and procedures, from such unstructured text is critical for enabling downstream applications like decision support and Computerized Physician Order Entry (CPOE) \cite{sutton2020overview, kuperman2003computer}. However, despite the adoption of CPOE systems, errors in order entry persist \cite{kinlay2021medication, campbell2006types}, and medication mistakes often arise during care transitions \cite{vira2006reconcilable}. This highlights the need for reliable methods to extract structured medical orders from clinical documentation.

To support such downstream tasks and reduce error rates, clinical information extraction (IE) methods have been developed to automatically identify entities and relations from free-text EHRs \cite{uzuner2010extracting, hahn2020medical}. These systems have enabled large-scale mining of clinical concepts for applications such as cohort identification, adverse event detection (ADE), and case surveillance \cite{sarmiento2016improving, landolsi2023information, ford2016extracting}. Within this domain, \textit{medical order extraction (MOE)} focuses specifically on identifying medical orders, such as medications, lab tests, or imaging, and structuring them into machine-readable formats \cite{xu2010medex}. Automating this process can reduce transcription burden, enhance care quality, and minimize errors in clinical workflows.

\begin{figure*}[ht]
\centering
\includegraphics[width=\textwidth, height=3.3in]{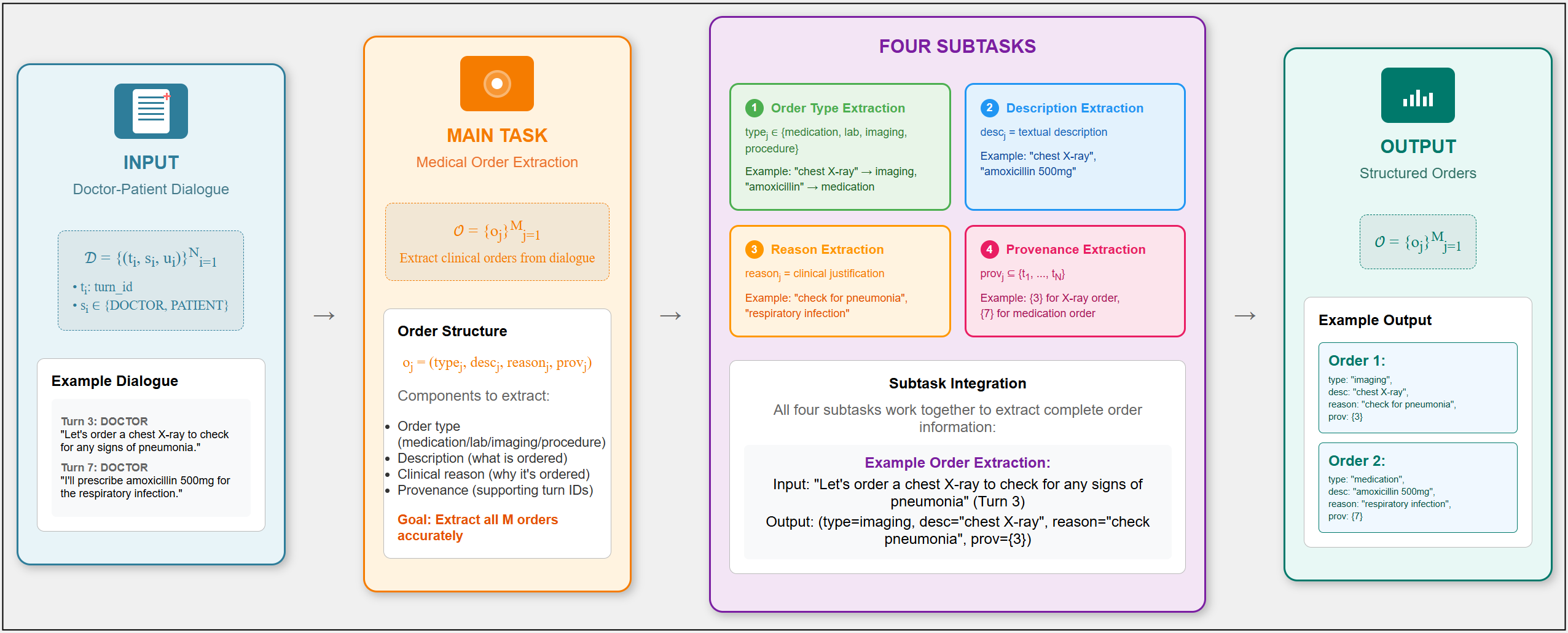} 
\caption{Overview of the MEDIQA-OE 2025 task: input transcripts, four subtasks (order type, description, reason, provenance), and the final structured output.}
\label{fig:task}
\end{figure*}

The \textbf{MEDIQA-OE 2025 Shared Task on Medical Order Extraction (OE)} \cite{MEDIQA-OE-Task} introduced a new benchmark to address this need. The task provides annotated multi-turn doctor–patient conversations and evaluates systems on their ability to extract structured medical orders, including medications, laboratory tests, imaging studies, and follow-up procedures, from conversational transcripts. In addition to identifying the order, systems must also extract the corresponding description and the reason or justification provided by the physician. This reflects real-world clinical documentation scenarios, where accurate interpretation of both the order and its rationale is essential.

In this paper, we describe our participation in the MEDIQA-OE task, which involves identifying and structuring various medical orders and their reasons with provenance grounding. Our approach uses \textbf{Meta’s LLaMA-4 Scout 17B} \cite{meta2025llama} model and relies on \textit{few-shot prompt engineering}, without any domain-specific fine-tuning or external knowledge sources. We curated exemplar prompts that capture conversational structures and medical order patterns. This approach allows us to evaluate the capabilities of general-purpose LLMs on domain-specific extraction tasks.

Our contributions are as follows:
\begin{itemize}
    \item We assess the feasibility and limitations of prompt-based approaches for structured information extraction in complex, safety-critical clinical domain tasks.
    \item We investigate the model's reasoning capabilities by analyzing how well it can identify the clinical justification (reason) for prescribed medications, lab tests, imaging, and follow-ups.
    \item We evaluate the ability of a non-medical, instruction-tuned LLM to perform medical order extraction from clinical text without any domain-specific fine-tuning. 
\end{itemize}

Our findings contribute to the growing body of work comparing general and domain-specific LLMs for clinical applications, highlighting prompting as a lightweight yet effective approach to structured prediction.

\section{Related Work}
The extraction of structured medical orders from unstructured clinical narratives has been a long-standing challenge in clinical Natural Language Processing (NLP), motivated by its potential to streamline clinical workflows, enhance decision support, and improve patient safety \cite{lussier2001automating,patrick2010high,uzuner2010extracting,uzuner20112010}. Early approaches were predominantly rule-based systems leveraging hand-crafted patterns, lexicons, and regular expressions to identify clinical entities and actions. Examples include systems built on platforms such as MedLEE \cite{friedman2000broad} and MetaMap \cite{aronson2010overview}, which mapped text spans to controlled vocabularies like the UMLS \cite{bodenreider2004unified}. These methods demonstrated high precision in restricted domains but suffered from limited transferability and scalability across institutions due to variations in clinical language and documentation styles.

The next generation of systems shifted toward statistical and machine learning approaches, which incorporated features from linguistic preprocessing (e.g., tokenization, POS tagging, dependency parsing) into classifiers such as Conditional Random Fields (CRFs) and Support Vector Machines (SVMs). Early examples in medication extraction, such as the 2009 i2b2 challenge systems \cite{patrick2010high}, demonstrated improved adaptability over purely rule-based methods, though their reliance on manually engineered features still posed challenges for transferability.

With the advent of deep learning, feature engineering was largely replaced by distributed representations learned directly from  \citeauthor{gan2023deep}. Recurrent Neural Networks (RNNs), particularly LSTMs \cite{hochreiter1997long} and BiLSTMs \cite{schuster1997bidirectional}, became popular for sequence labeling in clinical NLP, including medication extraction \cite{jagannatha2016structured,huang2015bidirectional,narayanan2022contextual,christopoulou2020adverse}. Attention mechanisms and hierarchical architectures further improved the capture of long-range dependencies, which is critical for modeling multi-turn dialogues and long EHR notes.

The introduction of transformer-based models \cite{vaswani2017attention} marked a significant leap in performance. Domain-specific transformers such as BioBERT \cite{lee2020biobert}, ClinicalBERT \cite{alsentzer-etal-2019-publicly}, and BlueBERT \cite{peng2019transfer} fine-tuned on biomedical corpora demonstrated substantial gains in extracting entities and relations from EHR data. These models leveraged self-attention to capture contextual relationships across long sequences, making them highly suitable for MOE from extended clinical narratives.

More recently, large language models (LLMs) such as GPT-3 \cite{brown2020language}, PaLM \cite{chowdhery2023palm}, and LLaMA \cite{touvron2023llama} have shown strong zero- and few-shot capabilities across domains, including clinical tasks. While most LLM work in healthcare has focused on general summarization, question answering, and entity recognition \cite{singhal2023large,moor2023foundation}, some studies have explored their application to structured MOE \cite{yang2020clinical,peng2023clinical,mahajan2023overview}. These approaches typically involve prompt engineering, in-context learning, or retrieval-augmented generation (RAG), sometimes without any domain-specific fine-tuning, to leverage LLMs’ reasoning and language understanding abilities.

\citeauthor{ford2016extracting,spasic2020clinical,grouin2023year} highlight a persistent gap between domain-specific models trained on narrow datasets and general-purpose LLMs that can adapt to new tasks with minimal supervision. While prior research has shown that LLMs trained on biomedical data can excel at extraction tasks, little is known about how non-biomedical, general-purpose LLMs perform in high-stakes domains like MOE when only minimal in-context examples are provided \cite{landolsi2023information,peng2023clinical}.

In this study, we position our work within the evolving trajectory from rule-based to LLM-based systems by focusing on the capabilities of a non-domain-specific model for the shared task. Our analysis complements prior work by quantifying how far prompt-engineered, general-purpose LLMs can go in structured clinical IE, without the cost and complexity of large-scale biomedical pretraining or fine-tuning, while identifying key strengths and weaknesses across different MOE subtasks.

\section{Task Description}

The MEDIQA-OE 2025 Shared Task \cite{MEDIQA-OE-Task} targets the extraction of structured medical orders from extended, doctor–patient conversations. The objective is to streamline clinical documentation, reduce provider workload, and ensure reliable capture of essential patient information from lengthy conversations.

Given a dialogue $\mathcal{D} = \{(t_i, s_i, u_i)\}_{i=1}^N$, where $t_i$ is the \texttt{turn\_id}, $s_i \in \{\texttt{DOCTOR}, \texttt{PATIENT}\}$ denotes the speaker, and $u_i$ is the utterance text, the goal is to predict a set of medical orders $\mathcal{O} = \{o_j\}_{j=1}^M$. 

Each medical order $o_j$ is divided into four subtasks as a tuple $(\texttt{type}_j, \texttt{desc}_j, \texttt{reason}_j, \texttt{prov}_j)$, where $\texttt{type}_j \in \{\texttt{medication}, \texttt{lab}, \texttt{imaging}, \texttt{followup}\}$, $\texttt{desc}_j$ is the textual description, $\texttt{reason}_j$ is the clinical justification, and $\texttt{prov}_j \subseteq \{t_1, \dots, t_N\}$ contains the supporting turn IDs.

Multiple medical orders may be present per dialogue, and systems must extract all relevant orders with accurate structure and provenance grounding. Success in this task requires models to handle long-range dependencies, differentiate between clinically relevant and incidental information, and produce outputs in a consistent, structured format that can be directly integrated into electronic health record (EHR) systems. Figure~\ref{fig:task} illustrates the input format, subtask definitions, and expected output.

\section{Dataset}

The MEDIQA-OE dataset \cite{MEDIQA-OE-2025-SIMORD-Dataset} consists of multi-turn doctor–patient conversations annotated with medical orders. Each instance is a JSON object containing an \texttt{id}, a list of \texttt{expected\_orders}, and a \texttt{transcript} of turns $(t_i, s_i, u_i)$, where $t_i$ is the turn ID, $s_i$ denotes the speaker, and $u_i$ is the utterance. Orders are annotated as $(\texttt{type}, \texttt{desc}, \texttt{reason}, \texttt{prov})$, with \texttt{prov} representing the supporting turn IDs.

\begin{table}[htpb]
\centering
\small
\begin{tabular}{lccccc}
\toprule
Set & \#Enc & Follow-Up & Imaging & Lab & Medication \\
\midrule
Train & 63 & 25 & 14 & 29 & 75 \\
Dev   & 100 & 41 & 26 & 71 & 117 \\
Test & 100 & - & - & - & - \\
\bottomrule
\end{tabular}
\caption{Number of encounters and order types per set. Gold labels for the test set have not been released.}
\label{tab:data-stats}
\end{table}

Transcripts were sourced from the PriMock57 \cite{papadopoulos-korfiatis-etal-2022-primock57} and ACI-Bench \cite{yim2023aci} datasets, with annotations merged using the official preprocessing script. The dataset, derived from the SIMORD corpus with an inter-annotator agreement of \textit{0.768}$\kappa$, was curated by experts following post-encounter documentation practices, capturing both explicit and implicit orders that often require multi-turn reasoning.

\paragraph{Dataset Analysis.}
Table~\ref{tab:data-stats} provides the distribution of the dataset with a breakdown of each \texttt{order\_type}. \textit{Medication }orders dominate in both training and development sets, followed by \textit{lab},\textit{ followup}, and \textit{imaging}, reflecting a clear class imbalance that may bias models toward frequent types in case of model fine-tuning. Dialogues are long, averaging $95.4$ turns in training, $102.1$ in development, and $101.6$ in test, with the longest spanning $290$ turns and over $2{,}900$ tokens, posing challenges for models with limited context windows. Across all sets, doctors produce the majority of content; for example, in the test set, they contribute $6{,}123$ turns and $89{,}449$ tokens compared to $4{,}037$ turns and $39{,}362$ tokens from patients. Follow-up suggestions appear in roughly one-third of encounters and align with annotated follow-up orders. Incomplete annotations are also present, with roughly one-fifth of orders lacking a \texttt{reason} field. \texttt{Provenance} spans averaging only $1$–$2$ turns, making both reason capture and evidence attribution challenging. 
These long contexts, skewed label distribution, implicit or missing reasons, and brief evidence spans, motivate models that can (i) maintain long-range dialogue state, (ii) generalize with little task-specific supervision, and (iii) ground outputs to cited turns. LLMs can handle extended inputs, adapt with few-shot prompts, and return structured fields with explicit provenance, making them the ideal candidate for this task.

\begin{figure}[htb]
\centering
\includegraphics[width=\columnwidth, height=5in]{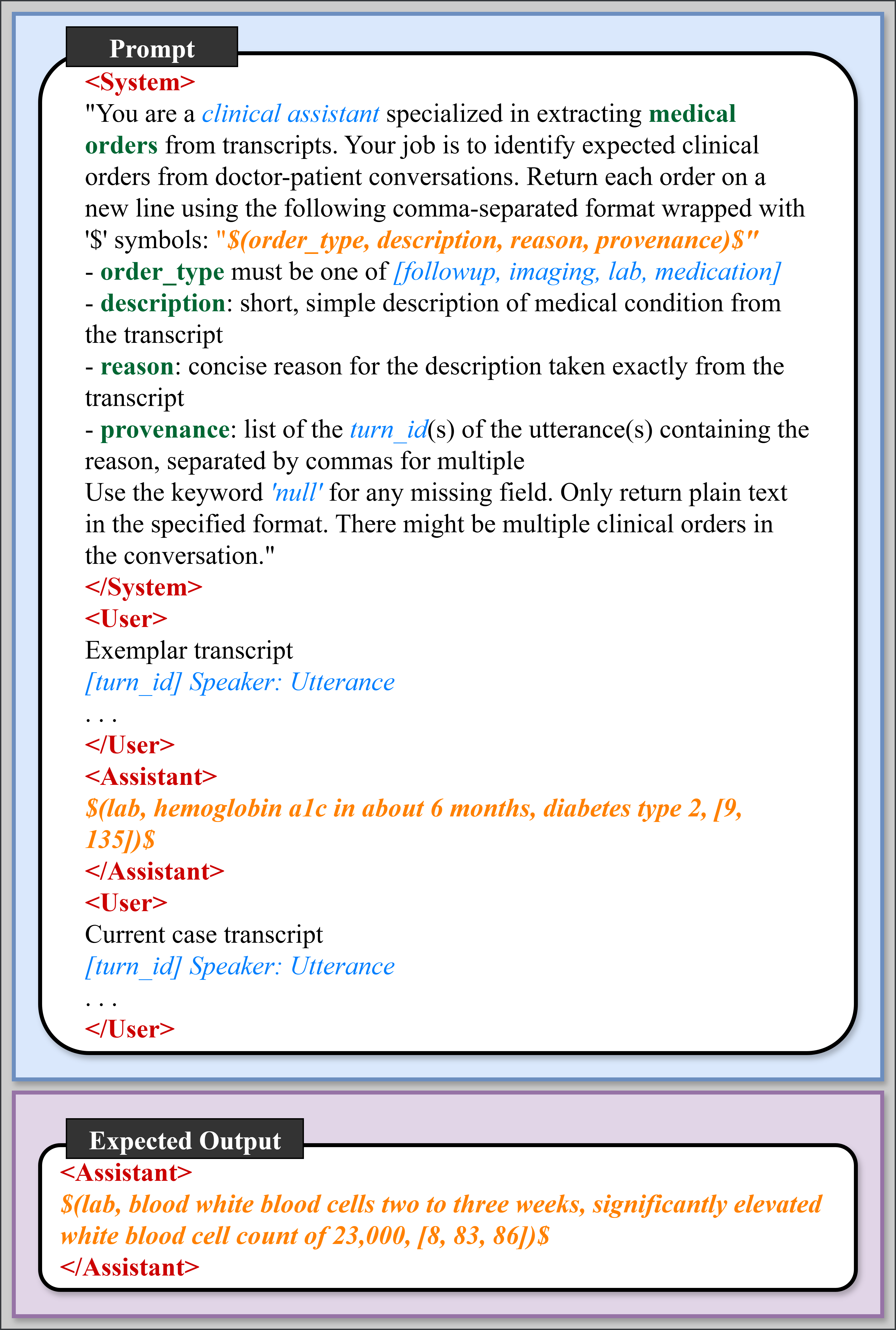} 
\caption{Few-shot prompt showing system instructions, exemplar input/output, the input query, and the expected model output.}
\label{fig:prompt}
\end{figure}

\section{Methodology}

We use several general-domain Meta-Llama \cite{touvron2023llama} models for MEDIQA-OE 2025, moving from zero-shot with a smaller model to few-shot with larger ones. We select Meta-Llama for its open weights, long context window, and strong instruction following ability, which is necessary for this task.

Input transcripts from the dataset were converted from JSON to a plain-text format with one utterance per line:
\[
\texttt{[turn\_id] Speaker: Utterance}
\]
This ensured that turn order and speaker roles were preserved for downstream reasoning.

\subsection{Model Configurations}
We evaluated three LLM configurations:
\paragraph{1. LLaMA 3 Inference (Zero-Shot):} First, \textit{meta-llama/}\textit{Llama-3-8B-Instruct} \cite{grattafiori2024llama} with no in-context examples. The system prompt defined the model’s role as a clinical assistant and specified the schema and constraints.
\paragraph{2. LLaMA-4 Inference (Zero-Shot):} Next, \textit{meta-llama/}\textit{Llama-4-Scout-17B-16E-Instruct} \cite{meta2025llama} with the same prompt design, leveraging a larger model for potentially better reasoning and grounding.
\paragraph{3. LLaMA-4 Inference (Few-Shot):} Finally, added a single in-context example from the training set, formatted as a user–assistant exchange preceding the inference case. The assistant’s example output illustrated the correct schema and provenance formatting, providing the model with a domain-specific reference.

\subsection{Prompt Design}
The prompt, illustrated in Figure~\ref{fig:prompt}, defined the model’s role as a clinical assistant tasked with extracting all medical orders from a doctor–patient conversation. It explicitly described the target output schema and the allowable values for \texttt{order\_type}, and it stated what each subtask should return. Additional guidance required \texttt{null} for missing fields, allowed multiple orders per conversation, and constrained \texttt{provenance} to turn IDs for the supporting utterances. The transcript was provided line by line with turn IDs and speaker roles, followed by either no example (zero-shot) or one exemplar with its gold output (few-shot). We reached this final prompt through iterative refinement: an initial version asked only for \texttt{order\_type}, \texttt{description}, \texttt{reason}, and \texttt{provenance} without role assignment, field definitions, or a fixed format, which led to long free-form text, difficult post-hoc parsing, and generic reasons that did not cite explicit spans. Assigning the clinical-assistant role and explaining each field improved grounding and produced more concise outputs. Requesting strict JSON next proved unreliable, as the model often added extra keys or commentary. We therefore switched to a comma-separated line format that the model followed more consistently. This process ensured clear instructions, a faithful mapping to the schema, and outputs that were both grounded and easy to parse.

\subsection{Post-Processing}
Following inference, raw model outputs were normalized and structured to match the required format. This process involved removing any extraneous text such as preambles or explanations, ensuring that all four fields were present, and explicitly assigning \texttt{null} to missing fields. The \texttt{order\_type} field was standardized to the allowed set, and the \texttt{provenance} field was validated to contain only integer turn IDs within the valid range for each conversation. Outputs were then serialized into JSON for evaluation. When predictions contained minor formatting issues, such as concatenated fields or misplaced delimiters, these were corrected automatically; predictions that could not be repaired were discarded to avoid evaluation errors.

\subsection{Experimental Setup}
All experiments were conducted on a High-Performance Computing (HPC) environment with NVIDIA A100 80GB GPUs using mixed-precision (\texttt{bfloat16}) inference to optimize memory usage and runtime. The maximum context length was set to 8{,}192 tokens with a generation limit of 1{,}024 tokens. Decoding parameters were fixed across runs (\texttt{temperature=0.2}, \texttt{top\_p=0.9}) to balance determinism and variability. Random seeds were fixed across the model, tokenizer, and generation routines for reproducibility.

\begin{table*}[h]
\centering
\begin{tabular}{lcccccc}
\hline
\textbf{Team Name} & \textbf{description} & \textbf{reason} & \textbf{order\_type} & \textbf{provenance} & \textbf{avg\_score} \\
\hline \hline
\textbf{MasonNLP} & 39.05 & 19.78 & 50.91 & 41.32 & 37.76 \\
\hline 
WangLab & 66.77 & 29.49 & 81.45 & 63.04 & 60.19 \\
silver\_shaw & 64.06 & 41.30 & 74.74 & 60.44 & 60.14 \\
MISo KeaneBeanz & 57.99 & 35.64 & 71.56 & 48.38 & 53.39 \\
EXL Health AI Lab & 54.45 & 30.50 & 66.17 & 52.47 & 50.90 \\
HerTrials & 19.61 & 8.99 & 29.59 & 5.61 & 15.95 \\
\hline
\end{tabular}
\caption{MEDIQA-OE 2025 leaderboard results (F1 in \%). Top six systems, rows sorted by average score; MasonNLP shown first for reference.}
\label{tab:leaderboard}
\end{table*}

\begin{table*}[h]
\centering
\begin{tabular}{lccccc}
\hline
\textbf{System} & \textbf{description} & \textbf{reason} & \textbf{order\_type} & \textbf{provenance} & \textbf{avg\_score} \\
\hline
LLaMA-3 8B (Zero-shot) & 30.20 & 13.95 & 40.79 & 27.10 & 28.01 \\
LLaMA-4 17B (Zero-shot) & 36.82 & 15.60 & 47.23 & 30.32 & 32.49 \\
LLaMA-4 17B (Few-shot) & \textbf{39.05} & \textbf{19.78} & \textbf{50.91} & \textbf{41.32} & \textbf{37.76} \\
\hline
\end{tabular}
\caption{Performance across experimental setups (F1 in \%). Best values are in bold.}
\label{tab:ablation}
\end{table*}

\subsection{Evaluation}
System outputs are first aligned to gold-standard orders through a description-based pairing process. Matching is performed on the \texttt{description} field after normalization, which lowercases text and removes selected punctuation. Orders are excluded from evaluation if they have an empty \texttt{description} or an \texttt{order\_type} outside the allowed set {\texttt{medication}, \texttt{lab}, \texttt{followup}, \texttt{imaging}}.

Once aligned, each field is scored with a metric suited to its content. The \texttt{description} field is evaluated with \textsc{Rouge-1} $F_1$, rewarding unigram overlap with the reference and granting partial credit for preserving key clinical terms even if phrasing differs. The \texttt{reason} field is also scored with \textsc{Rouge-1} $F_1$, capturing semantic similarity despite surface variation in justifications. The \texttt{order\_type} field uses a \textsc{Strict} $F_1$, counting only exact matches among the four permissible categories to penalize misclassification. The \texttt{provenance} field is evaluated with a \textsc{MultiLabel} $F_1$, treating provenance as a set of turn IDs and balancing precision (excluding unrelated turns) with recall (capturing all relevant turns).

The final shared-task score is the unweighted mean of the four primary field-level $F_1$ scores (\texttt{description\_ROUGE1\_f1}, \texttt{reason\_ROUGE1\_f1}, \texttt{order\_type\_Strict\_f1}, and \texttt{provenance\_MultiLabel\_f1}).

\section{Results and Discussion}
\label{sec:results}
\subsection{Leaderboard Performance}
The MEDIQA-OE 2025 shared task attracted participation from \textbf{17 teams}, producing a total of \textbf{105 submissions}. Our \textbf{MasonNLP} system, based on a few-shot prompting setup with the general-purpose LLaMA-4 17B model and no domain-specific fine-tuning, achieved an average $F_1$ score of \textbf{37.76}, placing competitively among the top-ranked systems. Table~\ref{tab:leaderboard} presents the top six leaderboard with subtask-specific scores, with our system listed first for clarity. Notably, this performance was obtained without incorporating clinical-domain pretraining or retrieval augmentation, competing against systems that leveraged specialized architectures or domain-specific resources.

\subsection{Ablation Study}
To better understand the impact of model scale and prompting strategy, we evaluated three configurations: LLaMA-3 8B zero-shot, LLaMA-4 17B zero-shot, and LLaMA-4 17B few-shot (final submission). Results in Table~\ref{tab:ablation} show a clear progression in average $F_1$ across configurations. Moving from LLaMA-3 to LLaMA-4 improved performance in all subtasks, especially \texttt{description} and \texttt{order\_type}, reflecting the larger model’s stronger capacity for identifying and categorizing medical orders in long transcripts. This aligns with the dataset’s high average turn count and doctor-heavy content, which demand robust long-context processing. Introducing a single in-context example further improved all four subtasks, with the largest relative gain in \texttt{provenance}, suggesting that even minimal task-specific guidance helps the model ground predictions more accurately and follow the required structured format.

\subsection{Discussion and Implications}
These findings confirm our initial hypothesis that larger, instruction-tuned LLMs provide measurable benefits for MOE from long, multi-turn dialogues, even without domain-specific fine-tuning. The improvements from the few-shot configuration validate our contribution, which claims that minimal in-context supervision can close part of the performance gap between general-purpose LLMs and domain-adapted systems. However, \texttt{reason} extraction remains the most challenging subtask, likely due to the implicit nature of many clinical justifications in the dataset. Similarly, while \texttt{provenance} accuracy improved, grounding still lags behind other subtasks, reflecting the difficulty of linking orders to scattered and sometimes indirect evidence in the dialogue.  

Overall, the results suggest that combining large general-purpose LLMs with carefully designed prompts and minimal in-context examples can yield competitive performance in structured clinical IE. Future gains may require integrating retrieval-based grounding or domain adaptation to better handle implicit reasoning and improve evidence alignment.

\section{Error Analysis}

Building on the results in Section~\ref{sec:results}, we conducted a detailed error analysis of our best-performing \texttt{LLaMA-4 17B} few-shot system to better understand its strengths and remaining challenges across the four subtasks. The development set offers gold-standard annotations for all fields, enabling both quantitative and qualitative assessment. The test set, lacking gold-standard annotations, is analyzed only for schema validity.

\begin{table}[ht]
\centering
\begin{tabular}{l c}
\hline
\textbf{Metric} & \textbf{Score} \\
\hline
\texttt{description\_ROUGE1\_f1} & 44.53 \\
\texttt{reason\_ROUGE1\_f1} & 25.13 \\
\texttt{order\_type\_Strict\_f1} & 57.28 \\
\texttt{provenance\_MultiLabel\_f1} & 40.17 \\
\hline
\texttt{avg\_score} & 41.78\\
\hline
\end{tabular}
\caption{Development set scores for the LLaMA-4 17B few-shot system (F1 in \%).}
\label{tab:dev-scores}
\end{table}

\begin{figure*}[ht]
\centering
\includegraphics[width=\textwidth]{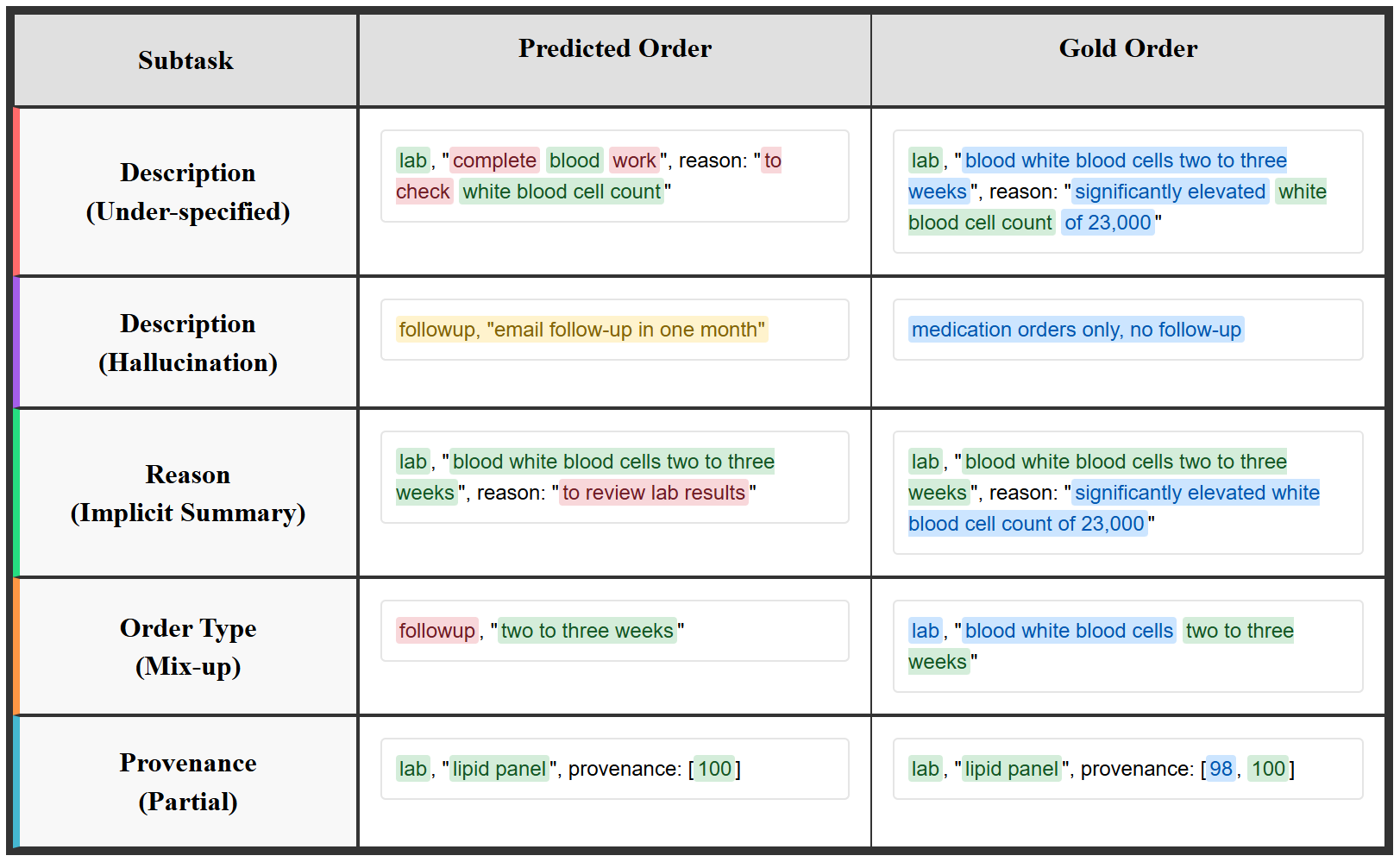} 
\caption{Examples of different error types for each subtask.}
\label{fig:error-examples}
\end{figure*}

\subsection{Development Set Analysis}
Table~\ref{tab:dev-scores} summarizes the official shared-task metrics for the development set. Consistent with leaderboard results, the model showed strong performance in \texttt{order\_type} classification and \texttt{description} extraction, while \texttt{reason} and \texttt{provenance} remained more challenging. The model also broke down a single order into multiple orders in some cases, as illustrated in Figure~\ref{fig:error-examples}. To explore why, we examined a few samples of matched and unmatched predictions, categorizing representative patterns for each subtask.

\paragraph{Description.}  
The model was able to identify the correct target of an order in most cases, even in multi-turn, context-heavy transcripts. Many predictions contained the correct general test or medication, but lacked finer details such as timing or exact test subtype. For example, \emph{``blood work''} was produced for the gold-standard \emph{``blood white blood cells two to three weeks''}. This indicates that the model successfully locates the core clinical action but sometimes omits modifiers, an area that could be enhanced by incorporating temporal and entity-specific cues.

\paragraph{Reason.}  
In most cases, the model provided a plausible reason aligned with the overall clinical context. For instance, it correctly linked a lab order to white blood cell count monitoring, though it occasionally summarized the reason more generally (\emph{``to review lab results''}) instead of including explicit values. This shows that the model is capable of long-context integration to capture the essence of clinical justification, with potential for refinement through methods that encourage inclusion of specific numeric and temporal evidence.

\paragraph{Order Type.}  
Order type classification was generally strong, but certain linguistic patterns led to confusion. Scheduling phrases (e.g., \emph{``two to three weeks''}) were sometimes interpreted as follow-up visits rather than scheduled labs. Invalid \texttt{order\_type}, present in 8 instances, included mentions of \textit{surgery (3), referral (1), and null\_type (4)}. Such mix-ups likely arise when multiple order-like actions occur in close proximity, and can be addressed by fine-tuning with examples emphasizing subtle category distinctions.

\paragraph{Provenance.}  
The model demonstrated the ability to identify at least one correct evidence turn for most orders, as in the case where it predicted provenance \texttt{[100]} while the gold-standard label included both \texttt{[98, 100]}. This partial grounding suggests that the model can reliably find the key confirmation turn, but may miss earlier reason turns when information is distributed. Expanding its retrieval capacity for dispersed evidence could close this gap.



\subsection{Test Set Analysis}
Out of all predicted orders, 20 (4.7\%) lacked a \texttt{description}, 8 (1.9\%) contained an invalid \texttt{order\_type} that included the same three keywords as we saw in the predicted orders of development set \textit{(surgery, referral, null\_type)}, 57 (13.3\%) were missing a \texttt{reason}, and 46 (10.8\%) omitted \texttt{provenance} identifiers. These results show that the system generally produces well-structured outputs with relatively few schema violations, though systematic omissions and field-level incompleteness directly reduce evaluation scores. Invalid \texttt{order\_type} predictions typically arose from ambiguous dialogue phrasing that led the model to select categories outside the permitted set {\texttt{medication}, \texttt{lab}, \texttt{followup}, \texttt{imaging}}. For \texttt{description}, beyond the 20 missing fields, 11 cases involved text not present in the transcript, reflecting hallucination or paraphrasing of plausible but unsupported orders. For \texttt{reason}, omission was dominant, with 57 missing values and 4 ungrounded justifications, indicating persistent difficulty in capturing implicit or distributed reason. For \texttt{provenance}, there are 46 missing spans. Partial grounding, common in the development set, likely persists here, underscoring the need for stronger evidence attribution.

\paragraph{Overall Observations.}  
The analysis shows that instruction-tuned LLMs, even without domain-specific fine-tuning, can handle complex, multi-turn clinical dialogues to extract actionable orders with reasonable accuracy. While finer details (e.g., exact timing, numeric values, dispersed evidence) are sometimes omitted, the model frequently identifies the correct order, reason, and at least one key supporting turn. Hallucinations, as with most LLMs, are still present, highlighting the potential benefit of RAG \cite{lewis2020retrieval}. With targeted enhancements, these strengths can be leveraged to develop robust clinical NLP systems capable of supporting real-world documentation workflows.

\section{Conclusion}

We addressed medical order extraction from multi-turn doctor–patient conversations using general-domain Meta-Llama models without domain-specific fine-tuning. The setup began with zero-shot prompting on a smaller model and then moved to few-shot prompting on larger models. A simple structured prompt returned order type, description, reason, and provenance with cited turns. The error analysis shows that the model struggles with temporal and numeric specificity, occasional hallucination, under-specific reasons, and partial provenance spans. These gaps narrowed with larger models and a few clear exemplars. The findings indicate that general domain LLMs are a viable base when guided by domain cues, retrieval to reduce hallucination, and schema validators for strict JSON. Overall, the study shows that instruction-tuned LLMs can handle long clinical dialogues with minimal adaptation and provides a practical template for grounded multi-field clinical IE with clear next steps on reason modeling, tighter provenance, and better balance across order types.

\section*{Limitations}
Our approach avoids any domain-specific pretraining or fine-tuning on clinical corpora. While integrating such specialization could potentially yield further gains, our goal was to assess the adaptability of a general-purpose, instruction-tuned LLM in a highly specialized medical order extraction task using only prompt engineering. This choice enables a fair evaluation of the model’s zero- and few-shot capabilities, providing insights into its out-of-the-box performance without reliance on costly domain-specific data or retraining. The strong results achieved by our few-shot LLaMA-4 system demonstrate that competitive baselines can be established under these conditions, laying the groundwork for future enhancements through targeted domain adaptation.


\bibliography{custom}

\appendix

\end{document}